\documentclass[12pt,journal, onecolumn, draftcls]{IEEEtran}
\usepackage{verbatim}
\usepackage{amsmath,graphicx}
\usepackage{mathrsfs}
\usepackage{amsopn}
\usepackage{cite}
\usepackage{amsthm}
\usepackage{epsfig,psfrag}
\usepackage{amsfonts,amssymb}
\usepackage{subfigure}
\usepackage{algorithm} 
\usepackage{algorithmic} 
\usepackage{graphics}
\usepackage{epsfig}
\usepackage{epstopdf}
\graphicspath{{ResultOne/}}
\usepackage{amsbsy}
\usepackage{tabularx}
\usepackage{amssymb} 
\usepackage{multirow} 
\usepackage{xcolor}
\usepackage{array}
\newcommand{\tabincell}[2]{\begin{tabular}{@{}#1@{}}#2\end{tabular}}

\begin{document}
\title{Image Fusion With Cosparse Analysis Operator}
\author{Rui~Gao,  {\it Student Member, IEEE},
Sergiy~A.~Vorobyov, {\it Senior Member, IEEE}, \\
and Hong~Zhao 
\IEEEaftertitletext{\vspace{0\baselineskip}}
\thanks{
R.~Gao is with Aalto University, Dept. Signal Processing and Acoustics, and with Northeastern University, Dept. Computer application technology, Shenyang 110819, China. (E-mail: rui.gao@aalto.fi). S.~A.~Vorobyov is with Aalto University, Dept. Signal Processing and Acoustics, FI-00076, AALTO, Finland. 
(E-mail: svor@ieee.org). H.~Zhao is with Northeastern University, 
Dept. Computer application technology, China. (E-mail: zhaoh@neusoft.com).
}}

\markboth{IEEE SIGNAL PROCESSING LETTERS}%
{Shell \MakeLowercase{\textit{et al.}}: Bare Demo of IEEEtran.cls
for Journals}
%
\maketitle
\begin{abstract}
The paper addresses the image fusion problem, where multiple images captured with different focus distances are to be combined into a higher quality all-in-focus image. Most current approaches for image fusion strongly rely on the unrealistic noise-free assumption used during the image acquisition, and then yield limited robustness in fusion processing. In our approach, we formulate the multi-focus image fusion problem in terms of an analysis sparse model, and simultaneously perform the restoration and fusion of multi-focus images. Based on this model, we propose an analysis operator learning, and define a novel fusion function to generate an all-in-focus image. Experimental evaluations confirm the effectiveness of the proposed fusion approach both visually and quantitatively, and show that our approach outperforms state-of-the-art fusion methods. 
\end{abstract}


\IEEEpeerreviewmaketitle

\section{Introduction}
\label{sec:intro}
Fusion of multi-focus images is a popular approach for generating an all-in-focus image with less artifacts and higher quality~\cite{Goshtasby2007fusion, wan2013multifocus}. It relies on the idea of combining a captured sequence of multi-focus images with different focal settings. It plays a crucial role in many fundamental fields such as machine vision, remote sensing and medical imaging~\cite{Zhang2016fusion, nencini2007remote, Cao2015fusion}. During the last decade, two types of fusion approaches have been developed: the transform domain-based approaches and the spatial frequency-based approaches. Most existing transform domain-based methods~\cite{burt1983laplacian, aslantas2010fusion, Li2008multi, zhou2014multi} work with a limited basis, and make fusion excessively dependent on the choice of a basis. The latter approaches~\cite{tian2012adaptive, Cunha2006fusion, rockinger1997image, zhang2009multifocus} require highly accurate sub-pixel or sub-region estimates, and thus, fail to perform well in elimination of undesirable artifacts.

A prevalent approach for image fusion is based on \emph{synthesis sparse model}. Various manifold fusion methods  have been proposed to explore this model~\cite{Zhang2016fusion, yang2010multifocusimage, Nejati2015Multi, rui2016coupled, ambat2010fusionalgorithms, li2015multi}. The core ideas here are to describe source images as linear combination of a few columns from a prespecified dictionary, merge sparse coefficients by a fusion function, and then generate an all-in-focus image using reconstructed sparse coefficients. While there has been extensive research on \emph{synthesis sparse model}, the \emph{analysis sparse model}~\cite{Elad2007analysis ,Dong2016analysis, Seibert2016analysis, Nam2013analysis} is a recent construction that stands as a powerful alternative. This new model represents a signal $\boldsymbol{x}$ by multiplying it with the so-called analysis operator $\boldsymbol{\Omega}$, and it emphasizes zero elements of the resulting analyzed vector $\boldsymbol{\Omega x}$ which describes the subspace containing the signal $\boldsymbol{x}$. It promotes strong linear dependencies between rows of $\boldsymbol{\Omega}$, leads to much richer union-of-subspaces, and shows the promise to be superior in various applications~\cite{Rubinstein2013analysis, Yaghoobi2013analysis, Hawe2013analysis}.

In this letter, we develop a novel fusion algorithm based on an analysis sparse model which allows for simultaneous restoration and fusion of multi-focus images. Specifically, we formulate the fusion problem as a regularized inverse-problem of estimating an all-in-focus image given its reconstructed form, and take advantages of the correlations among multiple captured images for fusion using a cosparsity prior. The corresponding algorithm exploits a combination of variable splitting and alternating direction method of multipliers (ADMM) to learn the analysis operator $\boldsymbol{\Omega}$, that promotes cosparsity. Furthermore, a fusion function generates the cosparse representation of an all-in-focus image. Advantages of this approach are: more flexible cosparse representation compared to the synthesis sparse approaches and better image restoration and fusion performance. The proposed approach also widens the applicability of analysis sparse model. 

\vspace{-4pt}
\section{Problem Formulation}
\label{sec:problem}
Let $\boldsymbol{I}_1, \boldsymbol{ I}_2,\cdots, \boldsymbol{ I}_K$ be a sequence of multi-focus images of the same scene acquired with different focal parameters. Our goal is to recover an all-in-focus image $\boldsymbol{I}_{\rm F}$ from $K$ captured images $\left\{\boldsymbol{ I}_k\right\}_{k=1}^K$. The model for describing the relationship between the sequence of multi-focus images $\left\{\boldsymbol{ I}_k\right\}_{k=1}^K$ and an unknown all-in-focus image $\boldsymbol{I}_{\rm F}$ can be formally expressed as 
\vspace{-8pt}
\begin{equation}\label{eq21}
\begin{split}
	\begin{aligned}
	\boldsymbol{I}_k = F_k (\boldsymbol{I}_{\rm F}) + \boldsymbol{V}, \qquad k = 1, \cdots , K
		\end{aligned}
	\end{split}
\end{equation}
where $F_k(\cdot)$ is a blurring operator that denotes the physical process of capturing $k$-th multi-focus image~\cite{Pertuz2013multi, Subbarao1993opt}, and 
$\boldsymbol{V}$ is an additive zero-mean white Gaussian noise matrix, with entries drawn at random from the normal distribution~$\mathcal{N}(0,{\sigma}^2)$.
The blurring operator $F_k(\cdot)$ in most cases is unknown and irreversible, therefore it is too complex to find a sequence $\left\{ F_k(\cdot) \right\}_{k=1}^K$ out of all possible operators. Instead, it is more favorable to seek a compromise between the physical modelling of image capture and signal approximation.

Assuming the cosparsity prior, each image patch $\boldsymbol{i}_{\rm F}\in\mathbb{R}^{m}$ of the image $\boldsymbol{I}_{\rm F}$ is said to have a cosparse representation over known analysis operator $\boldsymbol{\Omega}\in\mathbb{R}^{h\times m}$ with $h \geqslant m$, if there exits a sparse analyzed vector $\boldsymbol{\Omega}\boldsymbol{i}_{\rm F}\in\mathbb{R}^{h}$. The model emphasizes on $l$ zero coefficients of $\boldsymbol{\Omega}\boldsymbol{i}_{\rm F}$ and defines $\boldsymbol{\Omega}_{\Lambda}$ as a sub-matrix of $\boldsymbol{\Omega}$ with rows that belong to the co-support set $\Lambda$. The co-support set $\Lambda$ consists of the row indices, which determine the subspace that $\boldsymbol{i}_{\rm F}$ is orthogonal to. Then, $\boldsymbol{i}_{\rm F}$ can be characterized by its co-support. Every image patch $\boldsymbol{i}_{\rm F}$ is estimated by solving the following optimization problem 
\vspace{-4pt}
\begin{equation}\label{eq211}
\begin{split}
	\begin{aligned}
	\underset{\hat{\boldsymbol{i}}_{\rm F}} {\mathrm{min}} \left \| \hat{\boldsymbol{i}}_{\rm F} - \boldsymbol{i}_{\rm F} \right \|_2^2 \quad \mathrm{s.t. }\ \left \| \boldsymbol{\Omega} \hat{\boldsymbol{i}}_{\rm F} \right \|_1 \leqslant \epsilon,\ \boldsymbol{\Omega}_{\Lambda} \hat{\boldsymbol{i}}_{\rm F} = 0 
		\end{aligned}
	\end{split}
\end{equation}
where $\hat{\boldsymbol{i}}_{\rm F}$ is a denoised estimation of $\boldsymbol{i}_{\rm F}$, $\epsilon$ is a tolerance error, and $\| \cdot \|_1$ and  $\| \cdot \|_2$ are respectively the $l_1$-norm and $l_2$-norm of a vector. Since $\boldsymbol{i}_{\rm F}$ is unknown, we choose to use the cosparse representation vector $\boldsymbol{\Omega} \hat{\boldsymbol{i}}_{\rm F}$ for recovering $\hat{\boldsymbol{i}}_{\rm F}$ through optimal fusion of cosparse coefficients $\left\{\boldsymbol{\Omega} \boldsymbol{i}_k\right\}_{k=1}^K$. Using correlations among multiple images, the proposed approach defines fusion function as $T(\cdot)$ that generates an optimal cosparse representation and returns the corresponding indices of image patches.  
Thus, a natural generalization of the problem~\eqref{eq211} for recovering a clean all-in-focus image patch $\hat{\boldsymbol{i}}_{\rm F}$ is given as 
\begin{equation}\label{eq22}
\begin{split}
	\begin{aligned}
	\underset{k, \hat{\boldsymbol{i}}_{\rm F}} {\mathrm{min}} \left \| \hat{\boldsymbol{i}}_{\rm F}\! -\!\boldsymbol{i}_k \right \|_2^2 \quad 
	\mathrm{s.t. }\left \|T \!\left( \!\left\{ \boldsymbol{\Omega} \boldsymbol{i}_k \right\}_{k=1}^K \!\right) \right \|_1  \!\leqslant \epsilon,
	\boldsymbol{\Omega}_{\Lambda}\boldsymbol{i}_k = 0 \mathcal{.}
     \end{aligned}
	\end{split}
\end{equation} 
The role of $T (\cdot)$ in~\eqref{eq22} is to provide a meaningful constraint on how closely the optimal patch $\boldsymbol{i}_k$ approximates $\boldsymbol{i}_{\rm F}$. We replace the cosparse representation $\boldsymbol{\Omega} \hat{\boldsymbol{i}}_{\rm F}$ by the corresponding optimal $\boldsymbol{i}_k$ of the input image $\boldsymbol{I}_k$, with respect to the cosparse analysis operator $\boldsymbol{\Omega}$. Therefore, the major sub-problems here are learning the analysis operator $\boldsymbol{\Omega}$ and the fusion function~$T$.

\section{Robust Fusion via Analysis Sparse Model}
\label{sec:Proposed}
Analysis operator learning aims at constructing an operator $\boldsymbol{\Omega}$ suitable for a family of signals of interest. We first propose a practical approach for learning the analysis operator by variable splitting and ADMM. Then, with the analysis operator fixed, we define the optimal fusion function.

\vspace{-8pt}
\subsection{Analysis Operator Learning}
\label{learning}
Suppose a training set $\boldsymbol{Y} = [ \boldsymbol{y}_1\ \boldsymbol{y}_2\  \cdots \ \boldsymbol{y}_N] \in \mathbb{R}^{m \times N}$ is formed from a set of $N$ clean vectorized images $\boldsymbol{X} = [\boldsymbol{x}_1\ \boldsymbol{x}_2\  \cdots \ \boldsymbol{x}_N]\in \mathbb{R}^{m \times N}$ contaminated by an additive zero-mean white Gaussian noise $\boldsymbol{V} \in \mathbb{R}^{m \times N}$. Our task is to find $\boldsymbol{\Omega} \in \mathbb{R}^{h \times m}$ which enforces the coefficient vector $\boldsymbol{\Omega} \boldsymbol{x}_i$ to be sparse for each $\boldsymbol{x}_i$. This problem for $\boldsymbol{\Omega}$ can be cast as
\begin{equation}\label{eq31}
\begin{split}
\begin{aligned}
f ( \boldsymbol{\Omega}, \boldsymbol{x}_i) \triangleq \underset{ \boldsymbol{\Omega},  {\boldsymbol{x}_i}}{\mathrm{min}} \ \frac{1}{2} \left \| \boldsymbol{x}_i - \boldsymbol{y}_i \right\|_2^2 + \lambda \left \| \boldsymbol{\Omega} \boldsymbol{x}_i \right\|_1
\end{aligned}
\end{split}
\end{equation} 
\noindent where $\lambda$ is a regularization parameter. To prevent $\boldsymbol{\Omega}$ from being degenerate, it is common to constrain its rows $\left\{ \boldsymbol{\omega}_j \right \}_{j=1}^h$ so that they have their $l_2$-norms equal to one. Then, the constraint set can be described as
\begin{equation}\label{eq32}
\mathcal{C}\triangleq \left\{
\begin{aligned}
&\boldsymbol{\Omega}_{\Lambda_i} \boldsymbol{x}_i = 0, \quad \mathrm{rank} ( \boldsymbol{\Omega}_{\Lambda_i} ) = m - r, \; 1\leq i\leq N \\
&\left\|\boldsymbol{\omega}_j \right \|_2^2 = 1, \quad 1\leq j\leq h 
\end{aligned}
\right. 
\end{equation} 
where $\Lambda_i$ is the index set of the rows in $\boldsymbol{\Omega}$ corresponding to zero elements in $\boldsymbol{x}_i$, $\mathrm{rank} ( \cdot )$ denotes the rank of a matrix, and $m-r$ elements of $\boldsymbol{x}_i$ are zeros.  

The problem \eqref{eq31}-\eqref{eq32} is non-convex with respect to variables $\boldsymbol{\Omega}, \boldsymbol{x}_i$. A fundamental approach to addressing it is to alternate between the two sets of variables $\boldsymbol{\Omega}$ and $ \boldsymbol{x}_i$, i.e., minimizing over one while keeping the other fixed. Motivated by the first-order surrogate (FOS) approach~\cite{Mairal2015admm} and ADMM~\cite{Eckstein1992admm,Xie2012admm}, we propose the FOS-ADMM algorithm for cosparse coding. 

$\mathit{\ Cosparse\ coding:}$ With $\boldsymbol{\Omega}$ fixed, we update each column of $\boldsymbol{X}$. For notation simplicity, we drop the column index in $\boldsymbol{x}_i$ and $\boldsymbol{y}_i$. We observe that the objective function~\eqref{eq31} for fixed $\boldsymbol{\Omega}$, i.e., $f( \boldsymbol{x} )$ is the sum of two functions: $f_1 ( \boldsymbol{x} ) = \frac{1}{2} \left \| \boldsymbol{x} - \boldsymbol{y} \right\|_2^2$ and $f_2 ( \boldsymbol{v} ) = \lambda \left \| \boldsymbol{v} \right\|_1$ where $\boldsymbol{v} = \boldsymbol{\Omega} \boldsymbol{x}$. It enables us to transform the problem of minimizing $f(\boldsymbol{x})$ into the following constraint optimization problem
\begin{equation} \label{eq33}
\begin{split}
\begin{aligned}
\underset{\boldsymbol{x}, {\boldsymbol{v} }}{\mathrm{min}}\ f_1 (\boldsymbol{x}) + f_2 ( \boldsymbol{v} ) \quad \mathrm{s.t.}\ \   \boldsymbol{v} = \boldsymbol{\Omega} \boldsymbol{x} .
	\end{aligned}
	\end{split}
	\end{equation} 
The iterative algorithm to solve~\eqref{eq33} can be then expressed in the following ADMM~\cite{Eckstein1992admm} form
\begin{subequations}\label{34}
		\begin{align}\label{341}
		\boldsymbol{x}^{(t+1)}&=\mathrm{arg}\underset{\boldsymbol{x}}{\mathrm{min}}f_1(\boldsymbol{x})+
		\frac{\mu}{2}\left\| \boldsymbol{\Omega}\boldsymbol{x}-\boldsymbol{v}^{(t)}-\boldsymbol{d}^{(t)}\right\|_2^2\\ \label{342}
		\boldsymbol{v}^{(t+1)}&=\mathrm{arg}\underset{\boldsymbol{v}}{\mathrm{min}}f_2(\boldsymbol{v})+
		\frac{\mu}{2}\left\| \boldsymbol{\Omega}\boldsymbol{x}^{(t+1)}-\boldsymbol{v}-\boldsymbol{d}^{(t)}\right\|_2^2\\  
		\boldsymbol{d}^{(t+1)}&=\boldsymbol{d}^{(t)}-\left( \boldsymbol{\Omega}\boldsymbol{x}^{(t+1)}-\boldsymbol{v}^{(t+1)}\right)
		\label{343}
		\end{align}
	\end{subequations}
where $\mu$ is the augmented Lagrangian penalty and $\boldsymbol{d}^{(t)}$ is the vector of Lagrange multipliers at $t$-th iteration. Note that the updates for $\boldsymbol{x}$ and $\boldsymbol{v}$ are separated into~\eqref{341} and~\eqref{342}. The sub-problem~\eqref{341} is a convex quadratic problem, and it can be easily solved by the FOS approach~\cite{Mairal2015admm} that consists of solving  iteratively the optimization problem
\begin{equation}\label{eq35}
\begin{split}
	\begin{aligned}
	&\boldsymbol{x}=\mathrm{arg}\underset{\boldsymbol{x}}{\mathrm{min}}\ \mathcal{G}(\boldsymbol{x})
	\end{aligned}
	\end{split}
\end{equation} 
\noindent where $\mathcal{G} (\boldsymbol{x})\! =\! \left( \!{\triangledown_{\boldsymbol{x}} f_1 (\boldsymbol{x})} \!\right)^T ({\boldsymbol{x}^{(t+1)} \!-\! \boldsymbol{x}}) \!+\! \frac{1}{2} L\left \| \boldsymbol{x}^{(t+1)} - \boldsymbol{x} \right \|\!_2^2 $
+$ f_2(\!\boldsymbol{x})$, $\triangledown_{\boldsymbol{x}} f_1 (\boldsymbol{x})$ is the gradient of $f_1 (\boldsymbol{x})$, $L$ is the Lipschitz constant of $f_1 (\boldsymbol{x})$, and $( \cdot )^T$ stands for the transpose. 

As for the sub-problem~\eqref{342} of updating $\boldsymbol{v}$, it turns out to be a simple shrinkage problem. Thus, we just employ the soft-thresholding operator $\mathrm{soft} (\cdot)$, and the update \eqref{342} becomes
\vspace{-4pt}
\begin{equation}\label{eq36}
\begin{split}
	\begin{aligned}
	\!\!\!\!\boldsymbol{v}^{(\!t+1\!)}&\!=\! \mathrm{soft} \left( \boldsymbol{x}^{(t+1)}-\boldsymbol{d}^{(t)}, \frac{\lambda}{\mu}\right)\\
	&\!=\! \mathrm{sign}\!\left(\!\boldsymbol{x}^{(t+1)} - \boldsymbol{d}^{(t)}\!\right)\!
	\odot \!\mathrm{max} \!\left\{ \!\left| \boldsymbol{x}^{(\!t+1\!)}\!-\! \boldsymbol{d}^{(t)} \right|\! -\! \frac{\lambda}{\mu}, 0 \!\right\}
	\end{aligned}
	\end{split}
\end{equation} 
where $\mathrm{sign}(\cdot)$ is the sign function and $\odot$ stands for the component-wise product.
 
$\mathit{\ Analysis\ operator\ update:}$ With $\boldsymbol{X}$ fixed, we turn to updating $\boldsymbol{\Omega}$ that amounts to obtaining each row $\boldsymbol{\omega}_j$ of $\boldsymbol{\Omega}$. The update of $\boldsymbol{\omega}_j$ should be affected only by those columns of $\boldsymbol{X}$ that are orthogonal to it~\cite{Rubinstein2013analysis}. Denote ${J}$ as indices of those columns, then the corresponding optimization problem can be written as 
\begin{equation}\label{eq37}
\begin{split}
	\begin{aligned}
	 \underset{ {\boldsymbol{\omega}_j}} {\mathrm{min}} \sum_{i \in {J}} \left\| \boldsymbol{x}_i  - \boldsymbol{y}_i \right\|_2^2 \quad \mathrm{s.t.}\  &\boldsymbol{\Omega}_{\Lambda_i} \boldsymbol{x}_i=0, \  \left\|\boldsymbol{\omega}_j \right \|_2^2 = 1,\\
	&\mathrm{rank}(\boldsymbol{\Omega}_{\Lambda_i}) = \!m-\!r.
     \end{aligned}
	\end{split}
\end{equation} 
where $\{ \boldsymbol{x}_i \}_{i \in {J}}$, $\{ \boldsymbol{y}_i \}_{i \in {J}}$ form the sub-matrices of $\boldsymbol{X}$ and $\boldsymbol{Y}$ which containing ${J}$ columns found to be orthogonal to $\boldsymbol{\omega}_j$, respectively. 

The problem \eqref{eq37} leads to updates of the co-support sets ${\Lambda_i}$ in each iteration. Motivated by \cite{Rubinstein2013analysis}, we simplify the updates in ${\Lambda_i}$, and use the following approximation
\begin{equation}\label{eq38}
\begin{split}
	\begin{aligned}
 \underset{ {\boldsymbol{\omega}_j}}{\mathrm{min}} \sum_{i \in {J}} \left\| \boldsymbol{\omega}_j^T \boldsymbol{y}_i \right\|_2^2\quad\mathrm{s.t.}\  \left\|\boldsymbol{\omega}_j \right \|_2^2 = 1
     \end{aligned}
	\end{split}
\end{equation} 
as as an alternative which can be solved using the singular value decomposition (SVD) on the sub-matrix of $\boldsymbol{Y}$ formed by $\{ \boldsymbol{y}_i \}_{i \in {J}}$. 

\subsection{Local Optimal Fusion}
\label{fusion}
When the analysis operator $\boldsymbol{\Omega}$ is learned, we yet can not directly compute the cosparse representation of $\boldsymbol{\Omega}\boldsymbol{ i}_{\rm F}$. Instead, we work with the collection of the cosparse representations $\left\{\boldsymbol{\Omega}\boldsymbol{ i}_k\right\}_{k=1}^K$, and then seek the optimal one to recover the corresponding all-in-focus patch. 

Using the sliding window technique, each image $\boldsymbol{I}_k$ can be divided into small $n\times n$ patches, from left-top to right-bottom. For convenience, we introduce a matrix $\boldsymbol{W}_{i,j}$ to extract the $(i,j)$-th block from the image $\boldsymbol{I}_k$. Visible artifacts may occur on block boundaries, and we also introduce overlapping patch of length $p$ for each small patch, and demand that the reconstructed all-in-focus patches would agree to each other on the overlapping areas. According to \cite{yang2010multifocusimage}, the block, or equivalently, the set of indices $\{ i,\ j,\ k \}$ corresponding to the biggest value in the set $ \left\{ \left\| \boldsymbol{\Omega} \boldsymbol{W}_{i,j} \boldsymbol{I}_k \right\|_1\right\}_{k=1}^K$ is chosen to reconstruct the fused image. Thus, the problem of finding the optimal cosparse representation can be formulated as
\vspace{-2px}
\begin{equation}\label{eq39}
\begin{split}
	\begin{aligned}	
	&T ( \left\{ \boldsymbol{\Omega} \boldsymbol{W}_{i,j} \boldsymbol{I}_k \right\}_{k=1}^K) = \boldsymbol{\Omega} \boldsymbol{W}_{i,j} \boldsymbol{I}_{k} \\
	&\left\{ i,\ j,\ k \right\} = \mathrm{arg} \underset{i,j,k}{\mathrm{max}} \left( \left\{ \left\| \boldsymbol{\Omega} \boldsymbol{W}_{i,j} \boldsymbol{I}_k \right\|_1 \right\}_{k=1}^K \right) .
     \end{aligned}
	\end{split}
\end{equation} 

Given the optimal fusion function \eqref{eq39}, the fusion problem can be cast as the basis pursuit problem with the cosparse regularization term $\left \|\boldsymbol{\Omega} \boldsymbol{W}_{i,j} \boldsymbol{I}_k \right \|_1$. Thus, the problem \eqref{eq22} can be replaced by the following problem of finding the initial estimate $\hat{\boldsymbol{i}}_{{\rm F}_0}$ of the fused patch $\boldsymbol{i}_{F}$
\vspace{-4px}
\begin{equation}\label{eq40}
\begin{split}	
&\underset{\hat{\boldsymbol{i}}_{{\rm F}_0}} {\mathrm{min}}
	\sum_{i,j,k}\left \| \hat{\boldsymbol{i}}_{{\rm F}_0} -\boldsymbol{W}_{i,j}\boldsymbol{I}_k \right \|_2^2 \!\\
	&\mathrm{s.t. }\  \boldsymbol{\Omega}_{\Lambda} \boldsymbol{W}_{i,j}\boldsymbol{I}\!_k = 0,\
\sum_{i,j,k} \left \| \boldsymbol{\Omega} \boldsymbol{W}_{i,j} \boldsymbol{I}_k \right \|_1 \leq \epsilon   \mathcal{.}
	\end{split}
\end{equation}
Since the problem~\eqref{eq40} is convex, its solution can be efficiently computed by using many existing algorithms~\cite{Dong2016analysis, Rubinstein2013analysis, Yaghoobi2013analysis, Hawe2013analysis}. 

\vspace{-8px}
\subsection{Global Reconstruction}
\label{global}
The above explained local optimal fusion is used to recover local details for each all-in-focus patch, respecting spatial compatibility between neighbouring patches. In order to remove possible artifacts and improve spatial smoothness, the global reconstruction constraint between the initial image estimate $\hat{\boldsymbol{I}}_{{\rm F}_0}$, formed from all $\hat{\boldsymbol{i}}_{{\rm F}_0}$'s, and the final estimate $\hat{\boldsymbol{I}}_{\rm F}$ can be applied to make a further improvement. 

The size of $\boldsymbol{\Omega}$ is suitable to represent a small image patch, and it is too small to apply for the entire image. Therefore, we expand the size of $\boldsymbol{\Omega}$ and define 
\vspace{-4px}
\begin{equation}\label{eq41}
\begin{split}
 \boldsymbol{\Omega}_{\rm F} \triangleq
 \left[
 \begin{matrix}
   \boldsymbol{\Omega}\boldsymbol{W}_{1,1} &  \boldsymbol{\Omega}\boldsymbol{W}_{1,2} & \cdots &  \boldsymbol{\Omega}\boldsymbol{W}_{1,u}\\
   \boldsymbol{\Omega}\boldsymbol{W}_{2,1} &  \boldsymbol{\Omega}\boldsymbol{W}_{2,2}&  \cdots & \boldsymbol{\Omega}\boldsymbol{W}_{2,u} \\
   \vdots & \vdots &\cdots& \vdots\\
    \boldsymbol{\Omega}\boldsymbol{W}_{v, 1} &  \boldsymbol{\Omega}\boldsymbol{W}_{v,2}&  \cdots & \boldsymbol{\Omega}\boldsymbol{W}_{v,u}  
     \end{matrix} 
   \right]	
	\end{split}
\end{equation} 
as the global analysis operator where $u$ and $v$ are indices of the boundaries. 
Using the result from the local optimal fusion, the entire image can be redefined using the reconstruction constraint by solving the problem
\vspace{-2px}
\begin{equation}\label{eq42}
\begin{split}
&\underset{\hat{\boldsymbol{I}}_{{\rm F}}} {\mathrm{min}}
\left\| \hat{\boldsymbol{I}}_{\rm F} - \hat{\boldsymbol{I}}_{{\rm F}_0} \right \|_2^2 + \lambda^{\prime} \left\| \boldsymbol{\Omega}_{\rm F} \hat{\boldsymbol{I}}_{{\rm F}} \right\|_1
	\end{split}
\end{equation} 
where $\lambda^\prime$ is the parameter controlling the sparsity penalty and representation fidelity. Hence, the entire process of the optimal fusion is summarized in Algorithm~\ref{alg:fusion}.
\vspace{-5px}
\begin{algorithm}[h]
\caption{Fusion}
\begin{algorithmic}[1]
\label{alg:fusion}
\renewcommand{\algorithmicrequire}{\textbf{Input:}}
\renewcommand{\algorithmicensure}{\textbf{Output:}} 
\REQUIRE multi-focus images $\left\{\boldsymbol{I}_k\right\}_{k=1}^K$, analysis operator
$\boldsymbol{\Omega}$ obtained by analysis operator learning
\ENSURE the estimate of the all-in-focus image $\hat{\boldsymbol{I}}_{\rm F}$\
\FOR{each patch $\boldsymbol{i}_k$ from $\boldsymbol{I}_k$}
\STATE  Compute the fusing coefficients and generate the optimal set of $K$ indices of image patches, using ($\ref{eq39}$);\\
\STATE  Find the initial estimates $\hat{\boldsymbol{i}}_{{\rm F}_0}$, using ($\ref{eq40}$);\\
\ENDFOR
\STATE Define the global analysis operator $\boldsymbol{\Omega}_{\rm F}$, using ($\ref{eq41}$);\\ 
\STATE Reconstruct the all-in-focus image $\hat{\boldsymbol{I}}_{\rm F}$, using ($\ref{eq42}$).
\end{algorithmic}
\end{algorithm}

\vspace{-8px}
\section{Experimental Results}
\label{sec:Experimental}
We verify the restoration and fusion performance of the proposed approach by visual comparisons, and then discuss the quantitative assessments. We have tested our approach for a number of images, and here one representative example is shown. Specifically, fusion experiments over the standard multi-focus dataset \cite{Nejati2015fusion} are conducted. Throughout all the experiments, the tolerance error in the proposed approach is set as $\epsilon = 0.1$, the maximum number of iterations is $1000$, the patch size is $n=7$, and the overlapping length is $p=1$. During the analysis operator learning, the generated training set consists of $10,000$ two-dimensional normalized samples of size $7 \times 7$ extracted at random from the natural images. Considering the tradeoff between fusion quality and computations, the analysis operator size is fixed to $64 \times 49$. All the experiments are performed on a PC running Inter(R) Xeon(R) 3.40GHz CPU. 

\begin{figure*}[!htb]
\begin{minipage}{0.2\linewidth}
  \centerline{\includegraphics[width=3.55cm ]{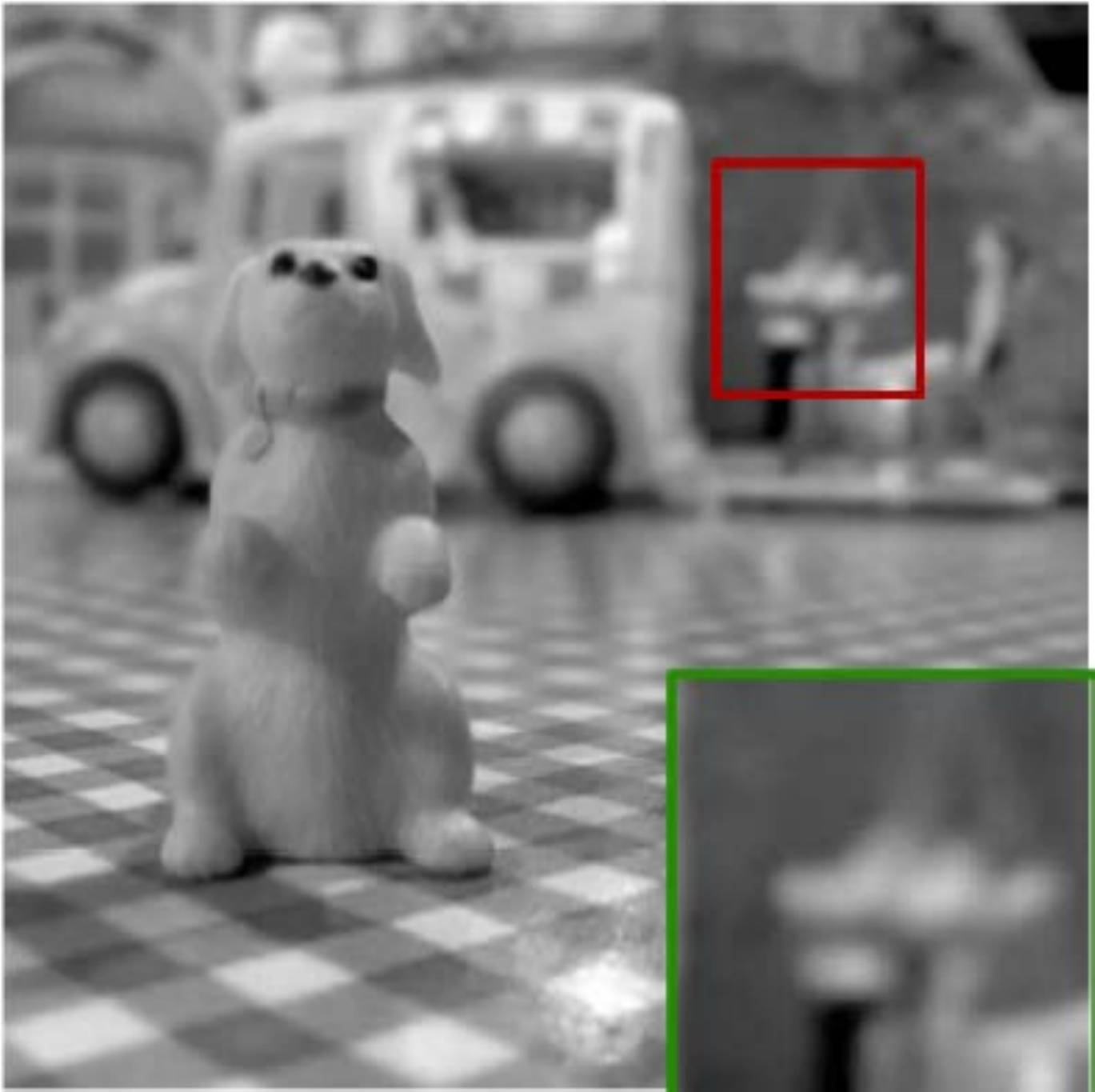}}
   \vspace*{-4pt}
  \centerline{(a)}
\end{minipage}%
\begin{minipage}{0.2\linewidth}
  \centerline{\includegraphics[width=3.55cm]{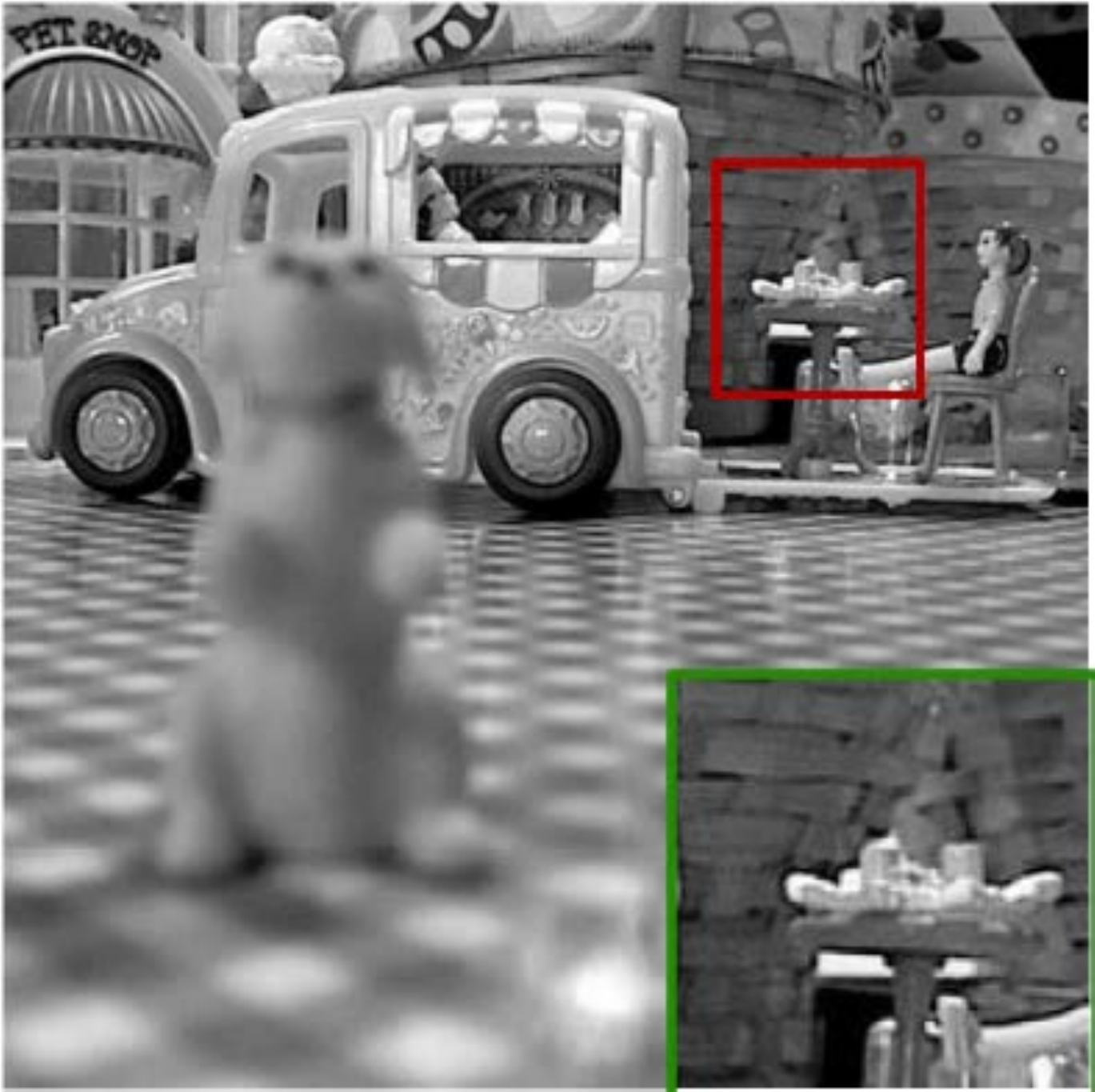}}
     \vspace*{-4pt}
  \centerline{(b)}
\end{minipage}%
\begin{minipage}{0.2\linewidth}
  \centerline{\includegraphics[width=3.55cm]{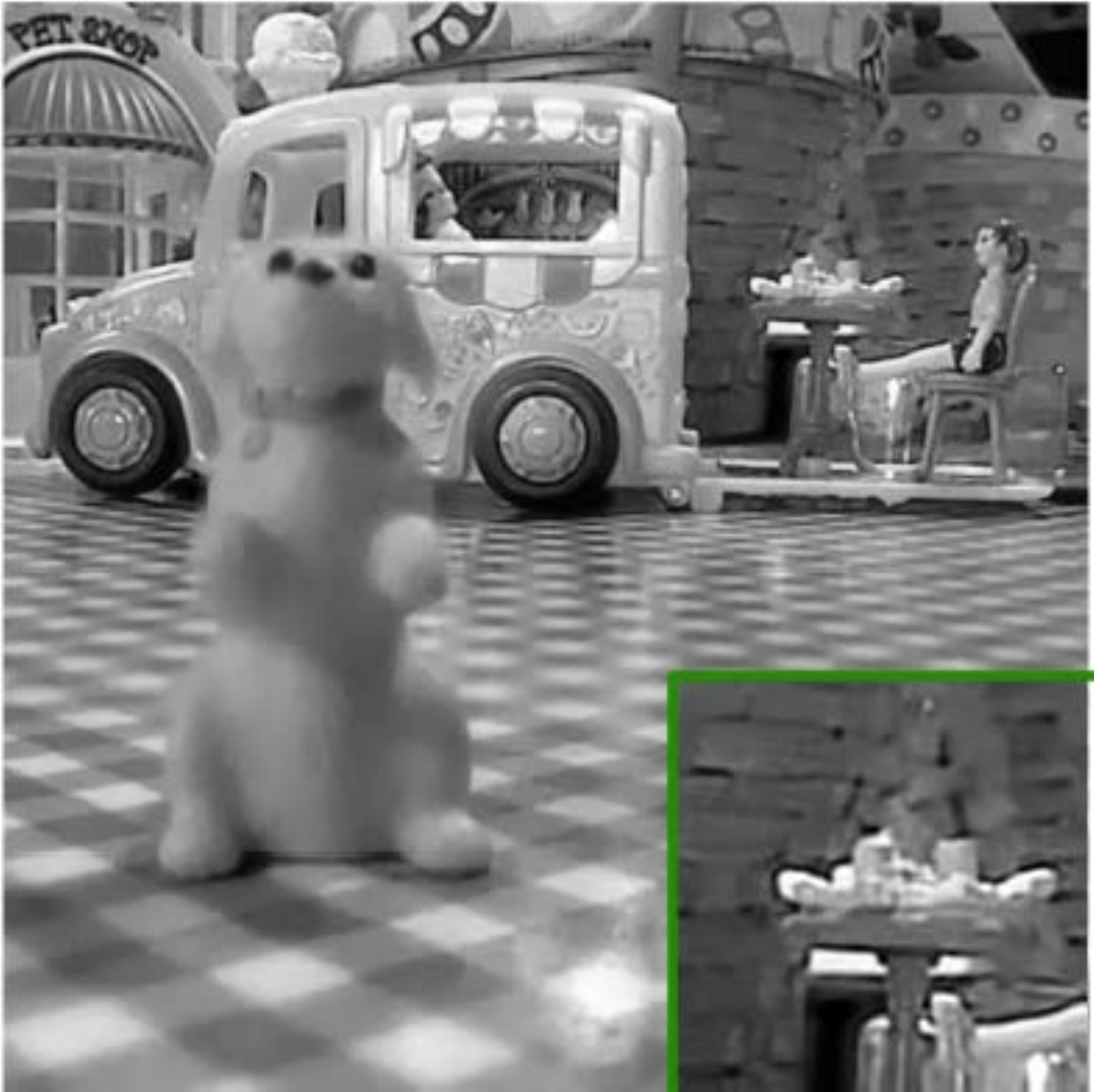}}
     \vspace*{-4pt}
  \centerline{(c)}
\end{minipage}%
\begin{minipage}{0.2\linewidth}
  \centerline{\includegraphics[width=3.55cm]{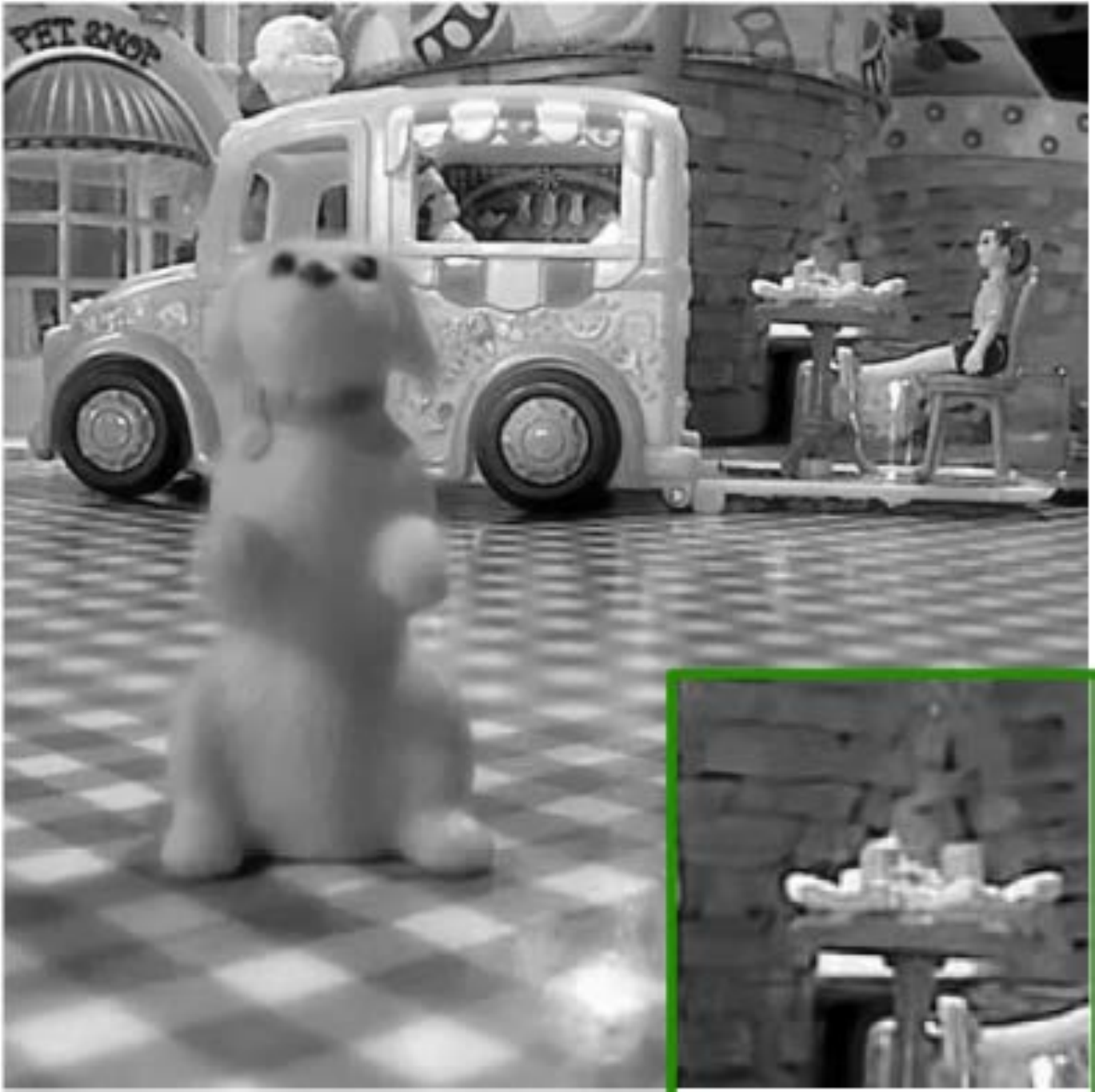}}
     \vspace*{-4pt}
  \centerline{(d)}
\end{minipage}%
\begin{minipage}{0.2\linewidth}
  \centerline{\includegraphics[width=3.55cm]{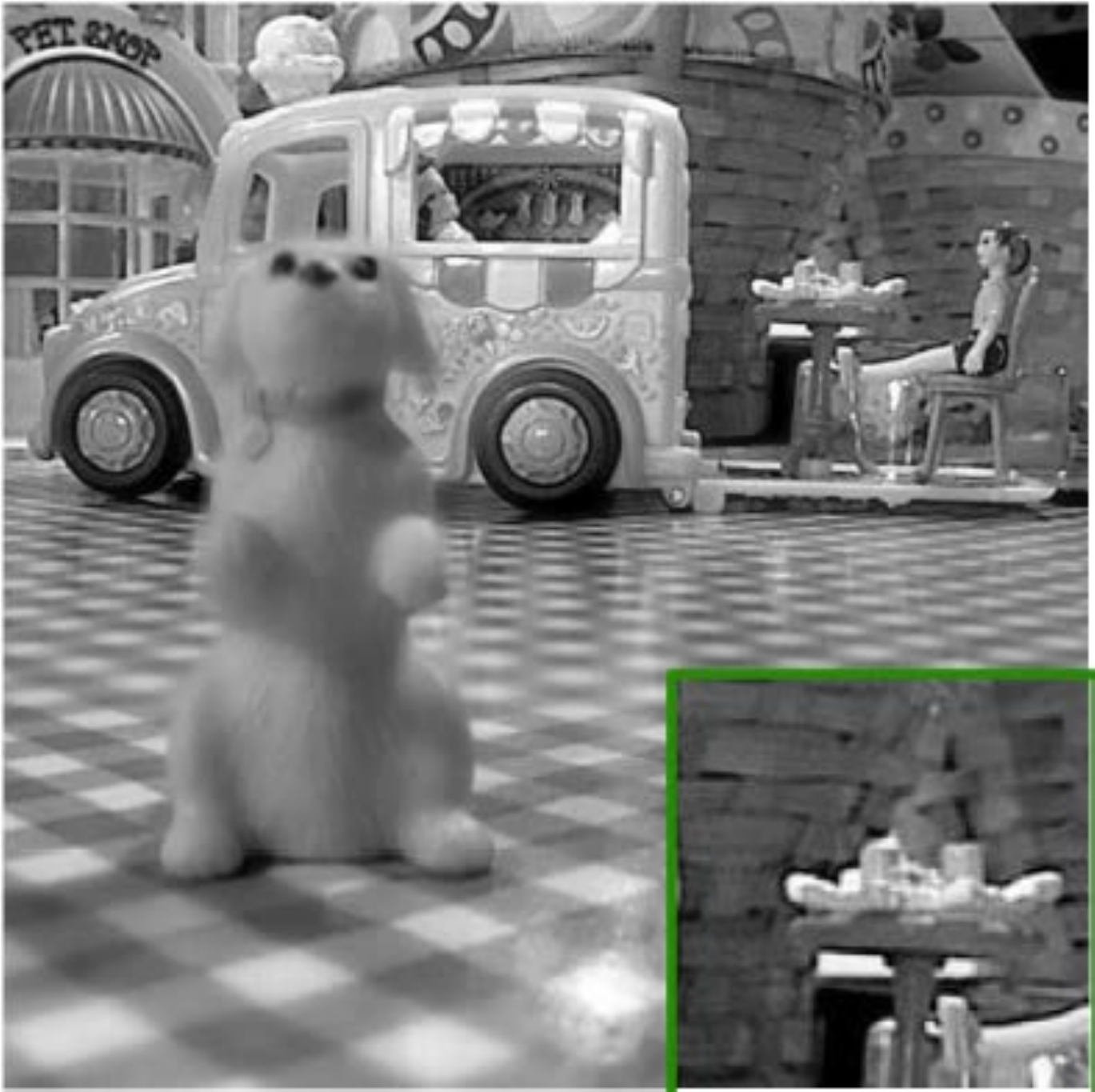}}
     \vspace*{-4pt}
  \centerline{(e)}
\end{minipage}%
\vfill
\centering
 \vspace*{-4pt}
\caption{Source images and the fusion results, with $\sigma=0$. (a) The first source image with focus on the front. (b) The second source image with focus on the background. Fused images obtained by SF-DCT (c), SR-KSVD (d), and the proposed method (e).}
\label{fig:disk_no}
\end{figure*}
\begin{figure*}[!htb]
\begin{minipage}{0.2\linewidth}
  \centerline{\includegraphics[width=3.55cm ]{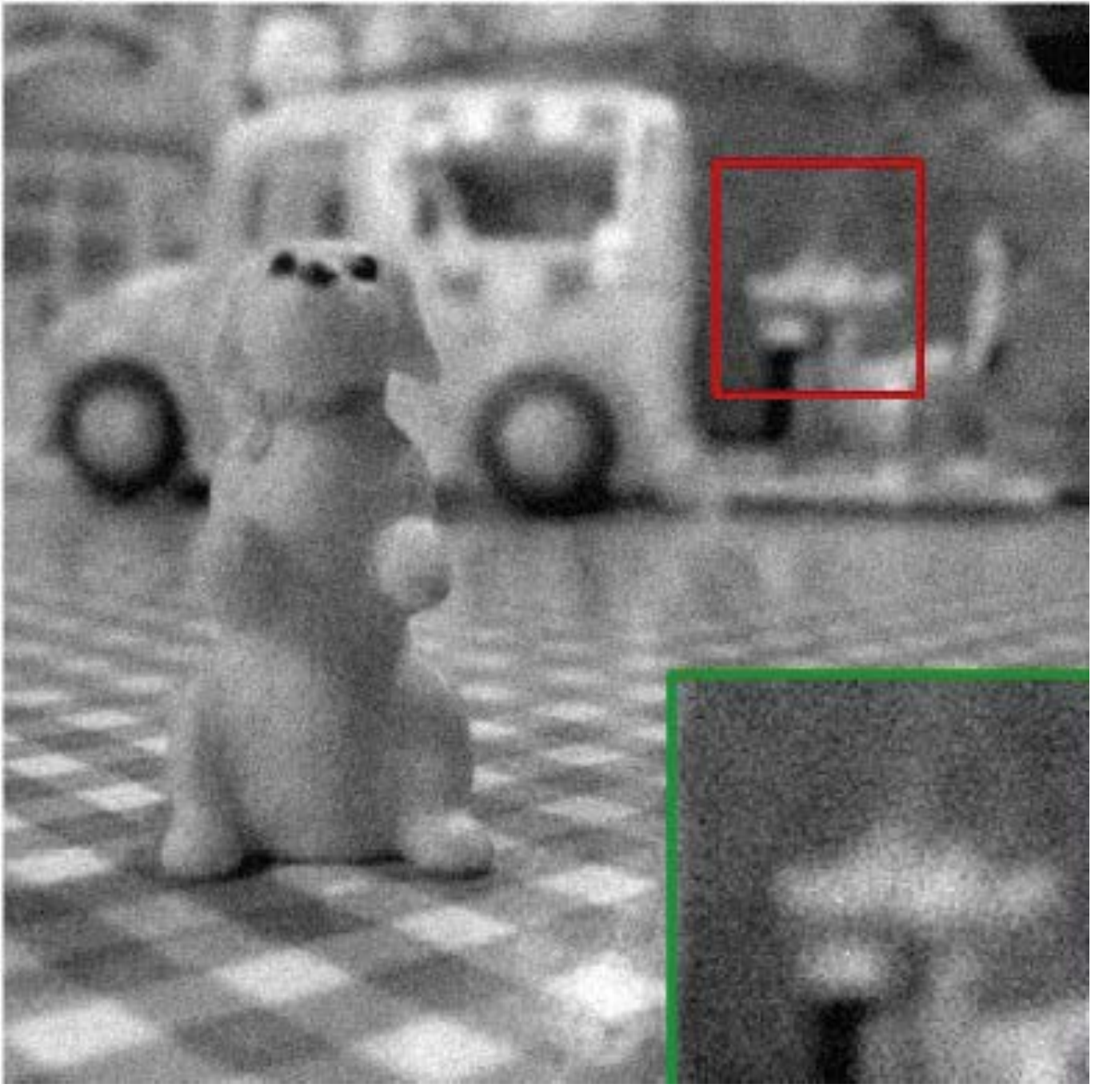}}
     \vspace*{-4pt}
  \centerline{(a)}
\end{minipage}%
\begin{minipage}{0.2\linewidth}
  \centerline{\includegraphics[width=3.55cm]{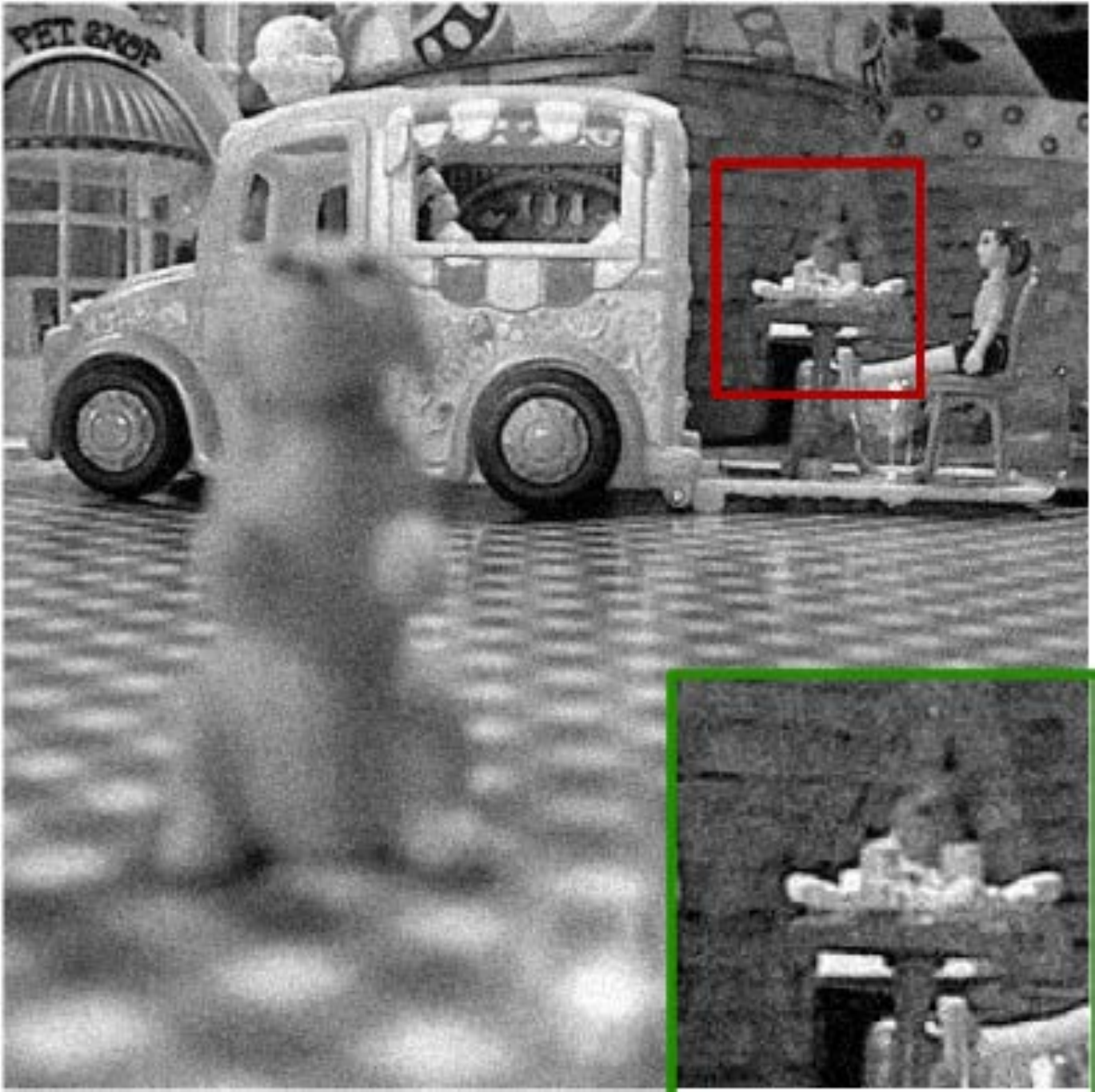}}
     \vspace*{-4pt}
  \centerline{(b)}
\end{minipage}%
\begin{minipage}{0.2\linewidth}
  \centerline{\includegraphics[width=3.55cm]{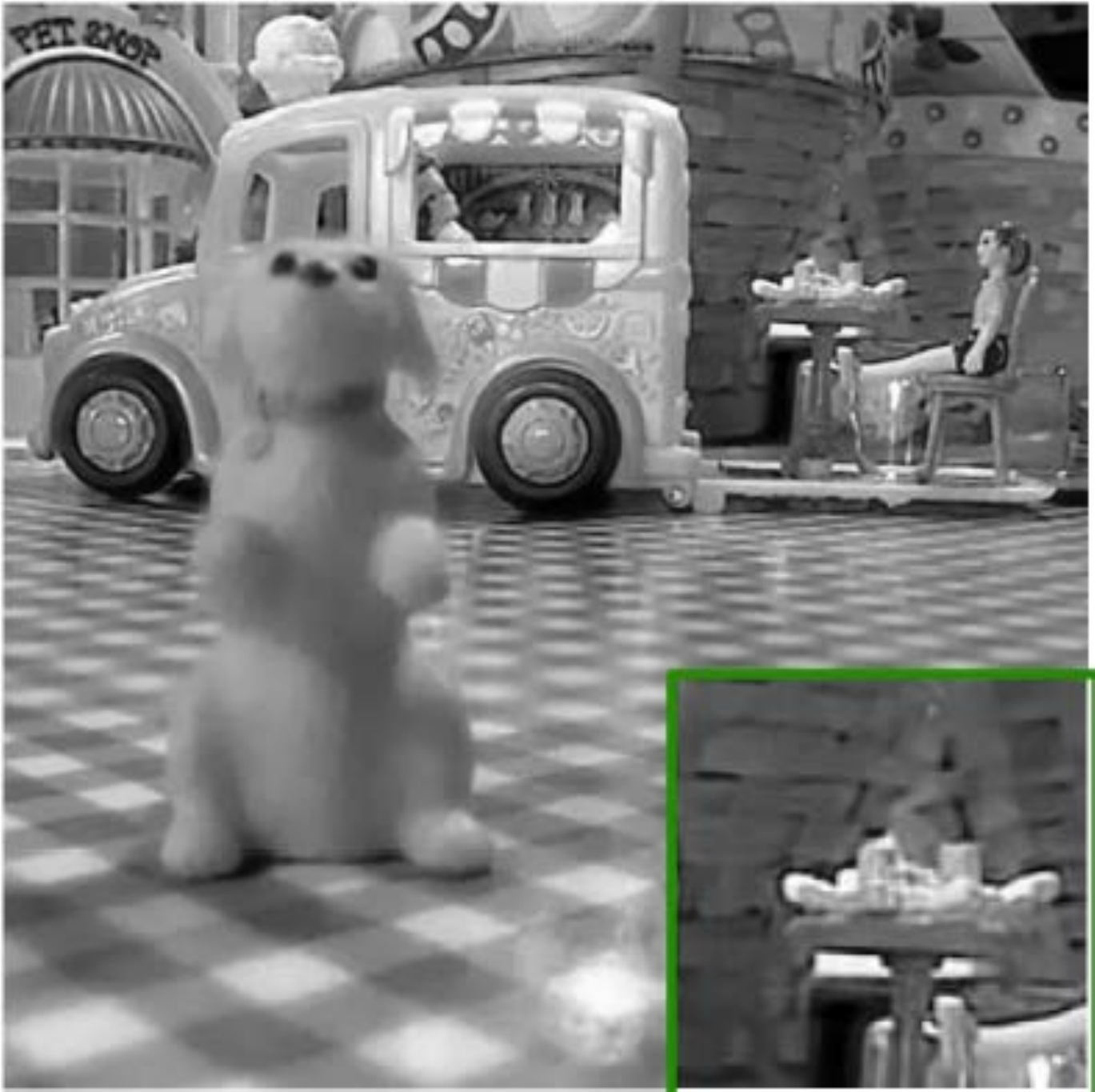}}
     \vspace*{-4pt}
  \centerline{(c)}
\end{minipage}%
\begin{minipage}{0.2\linewidth}
  \centerline{\includegraphics[width=3.55cm]{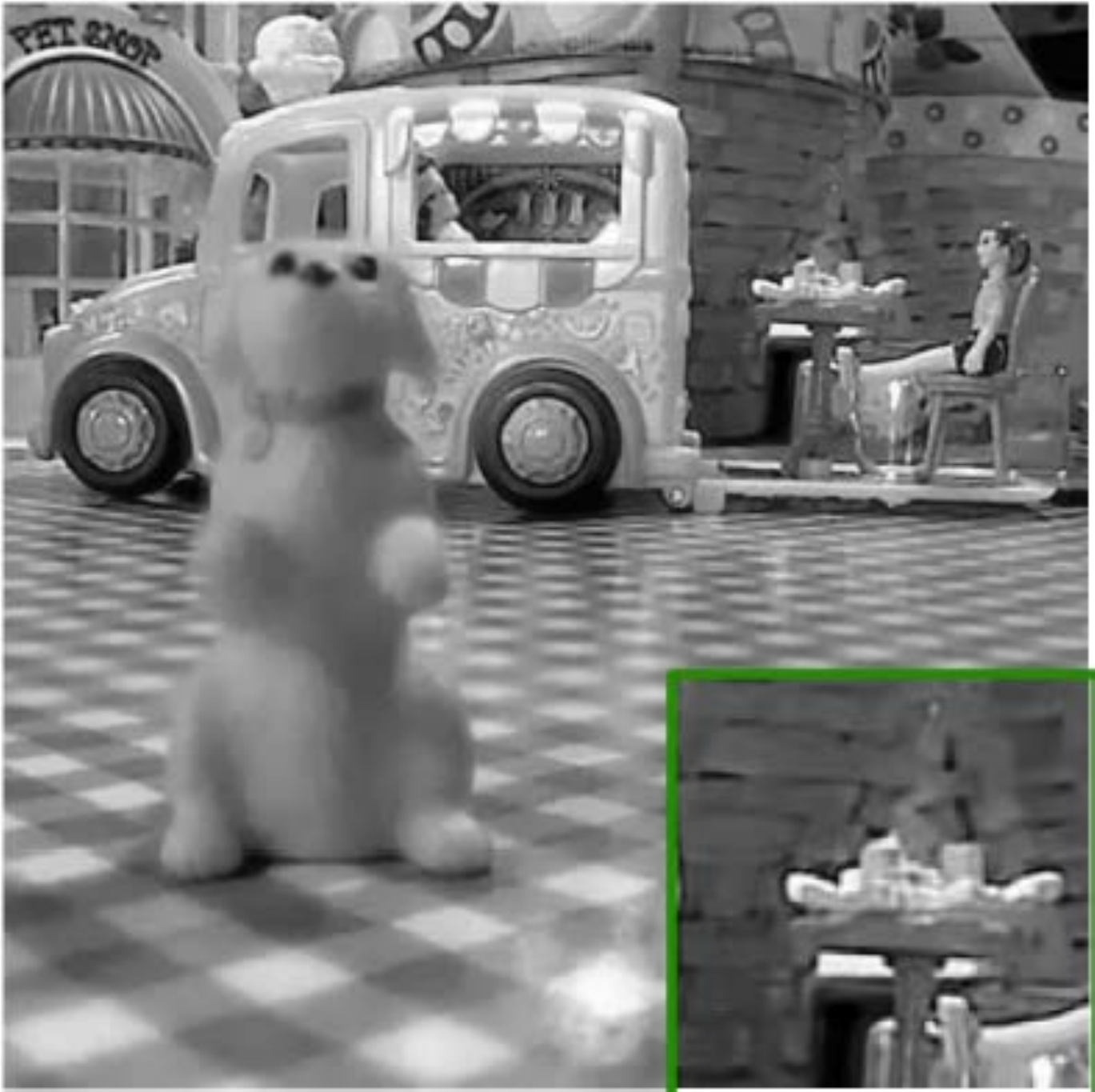}}
     \vspace*{-4pt}
  \centerline{(d)}
\end{minipage}%
\begin{minipage}{0.2\linewidth}
  \centerline{\includegraphics[width=3.55cm]{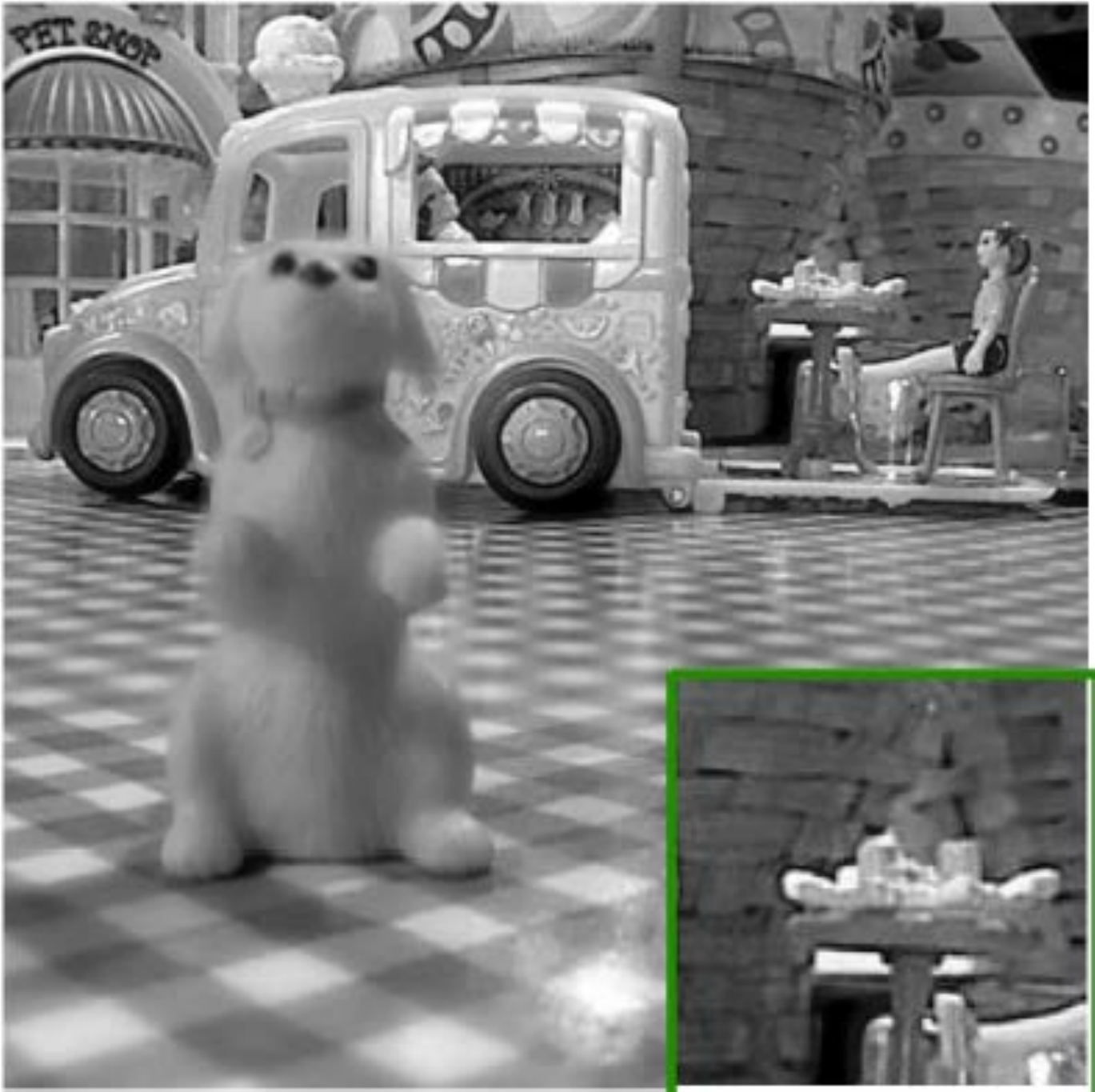}}
     \vspace*{-4pt}
  \centerline{(e)}
\end{minipage}%
\vfill
 \vspace*{-4pt}
\caption{Source images and the fusion results, with $\sigma=15$. Same order as in the Fig.~\ref{fig:disk_no}.}
\label{fig:disk_15}
\end{figure*}
\begin{figure}[!htb]  
\begin{minipage}{0.5\linewidth}
  \centerline{\includegraphics[width=3.3in,height=2.8in]{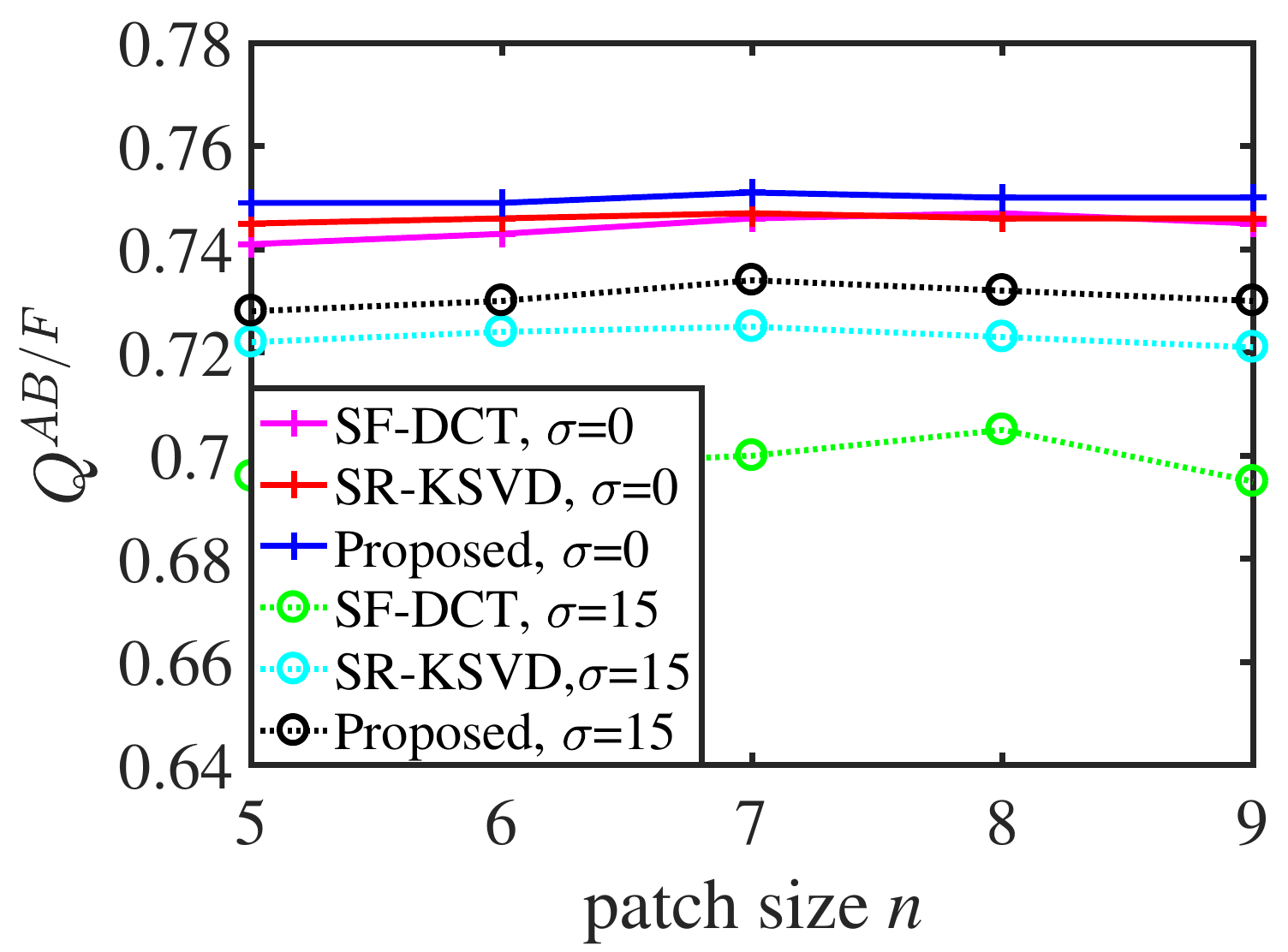}}
   \vspace*{-2pt}
  \centerline{(a)}
\end{minipage}%
\begin{minipage}{0.5\linewidth}
  \centerline{\includegraphics[width=3.3in,height=2.8in]{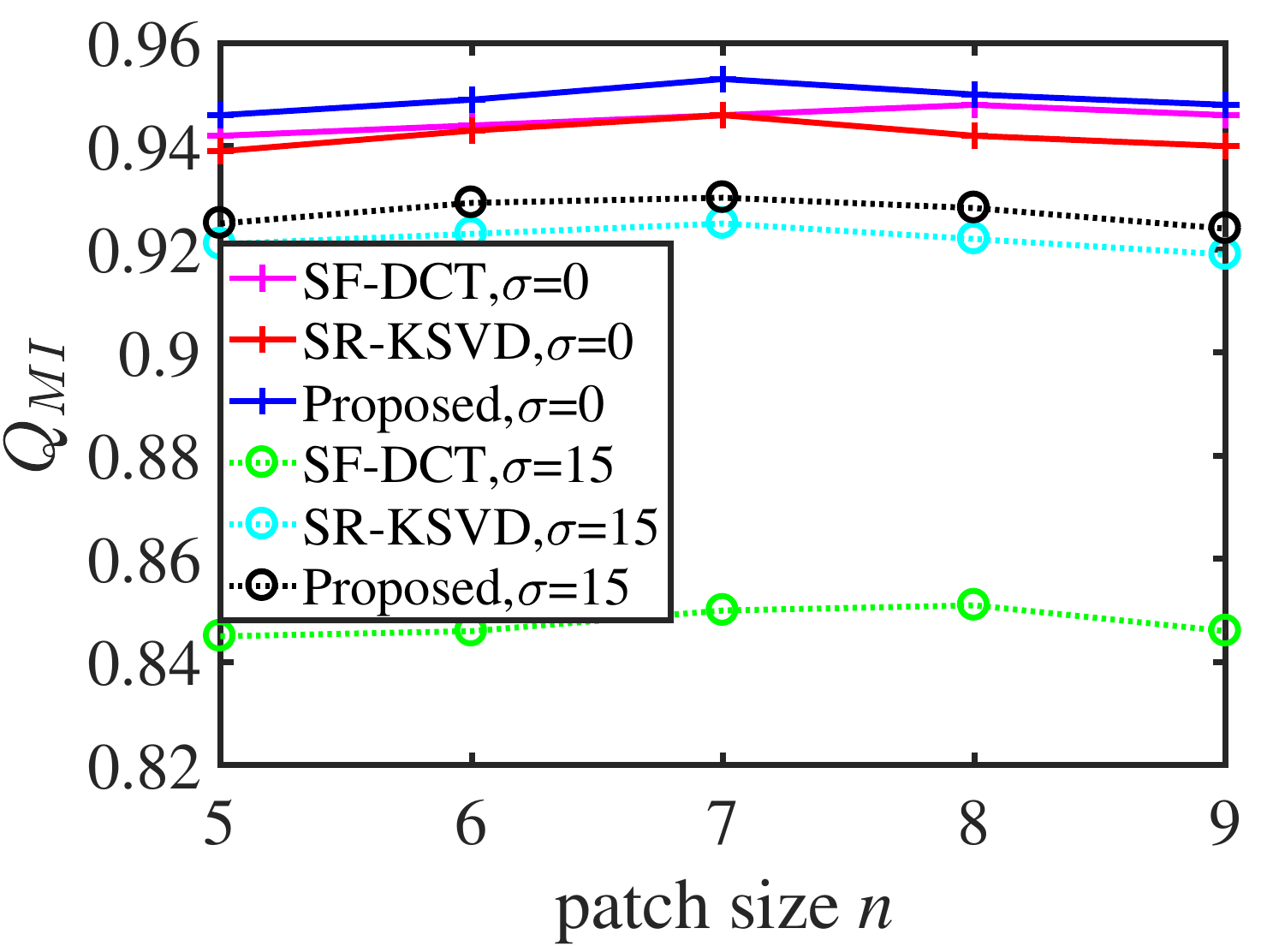}}
   \vspace*{-2pt}
  \centerline{(b)}
\end{minipage}%
\vfill
\centering
 \vspace*{-4pt}
\caption{Fusion performances versus patch size $n$.}
\label{fig:n}
\end{figure}
\begin{figure}[!htb]  
\begin{minipage}{0.5\linewidth}
  \centerline{\includegraphics[width=3.3in,height=2.8in]{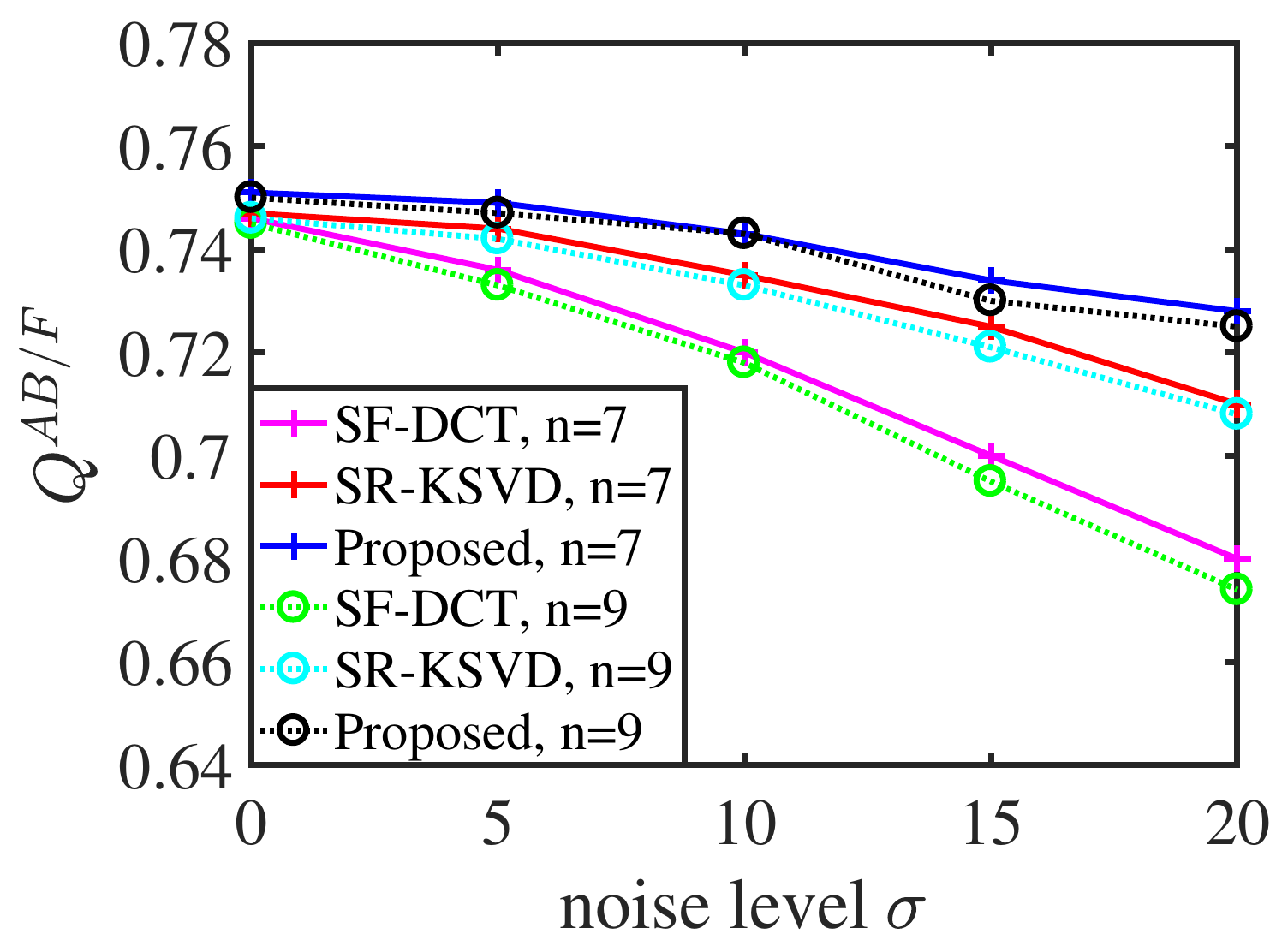}}
   \vspace*{-2pt}
  \centerline{(a)}
\end{minipage}%
\begin{minipage}{0.5\linewidth}
  \centerline{\includegraphics[width=3.3in,height=2.8in]{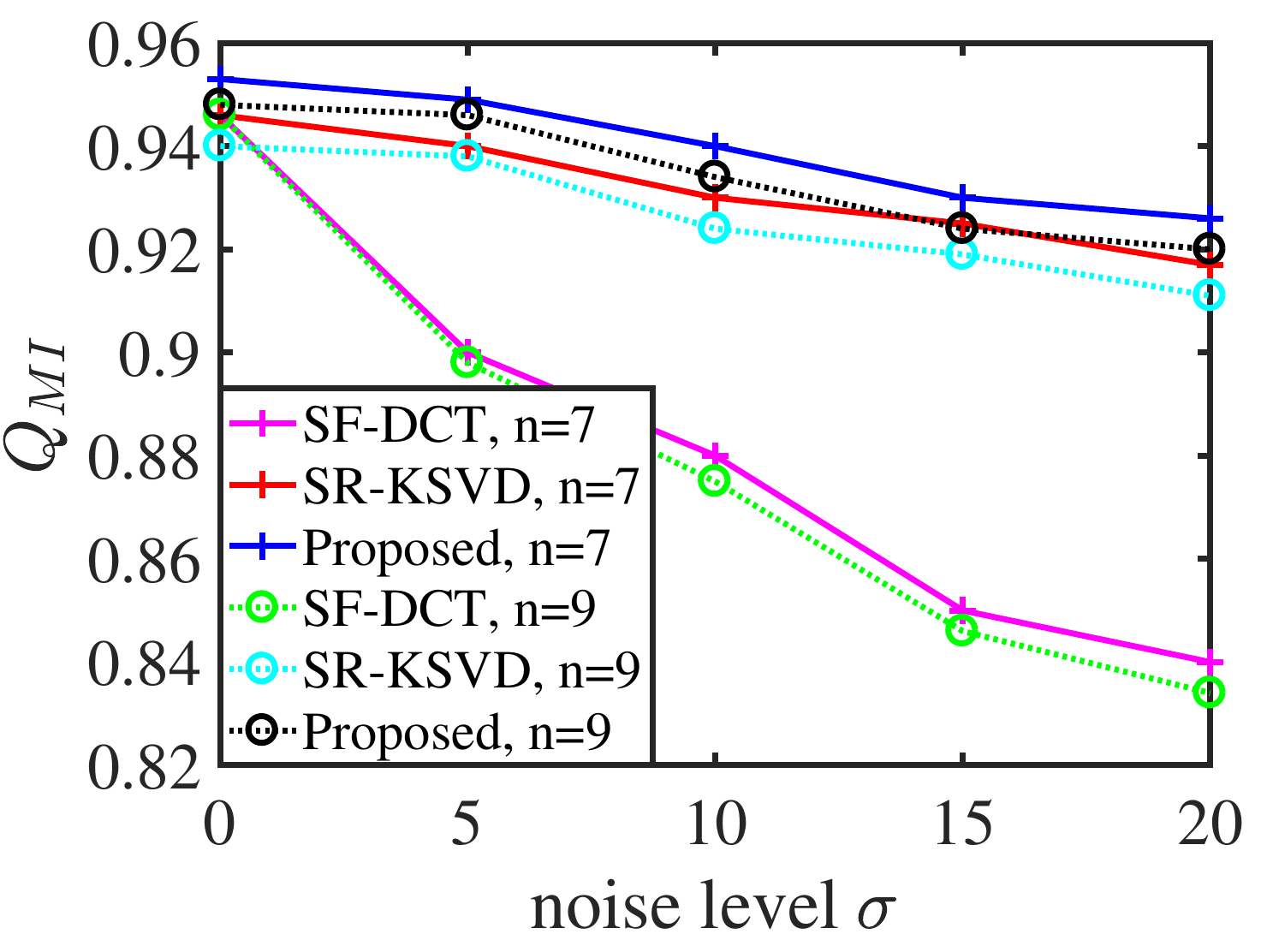}}
 \vspace*{-4pt}
  \centerline{(b)}
\end{minipage}%
\vfill
\centering
 \vspace*{-4pt}
\caption{Fusion performances versus noise level~$\sigma$.}
\label{fig:sigma}
\end{figure}
In the noise-free ($\sigma=0$) and noisy ($\sigma=15$) cases, the proposed approach is compared with well-known fusion approaches, including the image fusion approach based on spatial
frequency in discrete cosine transform (\mbox{SF-DCT})~\cite{Cao2015fusion} and the sparse representation K-SVD-based image fusion approach (\mbox{SR-KSVD})~\cite{yang2010multifocusimage}. 
The fusion results of the noise-free images ``Dog'' are shown in Fig.~\ref{fig:disk_no}, including the magnified details in the lower right corners of the images. There are noticeable differences in the edge of the wall. The \mbox{SF-DCT} method (see Fig.~\ref{fig:disk_no}(c)) produces blocking artifacts, and the \mbox{SR-KSVD} method (see Fig.~\ref{fig:disk_no}(d)) introduces undesired smoothing. Our proposed method (see Fig.~\ref{fig:disk_no}(e)) eliminates some artificial distortions, and gives better visual result. To test the robustness of our approach, we add Gaussian noise to the multi-focus images. In Fig.~\ref{fig:disk_15}, the results for the approach tested are shown for $\sigma=15$. Note that the \mbox{SF-DCT} method needs the denosing preprocessing, and then fuses multi-focus images.
Fig.~\ref{fig:disk_15}(c) shows circle blurring effect around strong boundaries. The image (see Fig.~\ref{fig:disk_15}(d)) also shows the blurring effect for the SR-KSVD method. Our approach is capable of providing restoration and fusion simultaneously, and it performs the best as it visually appears in Fig.~\ref{fig:disk_15}(e).

More objectively, we test the impact of different parameters selection on the proposed approach. The objective evaluation is based on the following two state-of-the-art fusion performance metrics: $Q_{MI}$~\cite{hossny2008comments}, which measures how well the mutual information from the source images is preserved in the fused image; and $Q^{AB/F}$~\cite{xydeas2000objective}, which evaluates how well the edge information transfers from the source images to the fused image. The values of $Q_{MI}$ and $Q^{AB/F}$ range from $0$ to $1$, with $1$ representing the ideal fusion. First, we conduct several experiments for different patch sizes, and compare the performance in the noise-free ($\sigma=0$) and noisy ($\sigma=15$) cases for the aforementioned methods in Fig.~\ref{fig:n}. The employed patch sizes are $n = \left\{ 5,6,7,8,9\right\}$. In either case $\sigma = \{ 0, 15 \}$, the values of $Q^{AB/F}$ (see Fig.~\ref{fig:n}(a)) and $Q_{MI}$(see Fig.~\ref{fig:n}(b)) for the proposed approach are always larger than for the \mbox{SF-DCT} and \mbox{SR-KSVD} methods. It means that our approach preserves well the mutual information and transfers efficiently the edge information from source images. When patch size is $7 \times 7$, the values of $Q^{AB/F}$ and $Q_{MI}$ are optimal. Thus, we set the patch size $n=7$, and also conduct fusion experiments with different noise levels $\sigma = \left\{0, 5, 10, 15, 20 \right\}$. The results are shown in Fig.~\ref{fig:sigma}. It can be seen that all the methods tested show larger values when $\sigma$ is equal to zero. With the increase of the noise level, the values of $Q_{MI}$ and $Q^{AB/F}$ gradually decrease, while the proposed method performs the best. Table~\ref{table:1} presents the average running time of the aforementioned methods. 
As expected, the proposed approach achieves the restoration and fusion with high-quality in reasonable time.

\begin{table}[h]
\scriptsize
\caption{Time measure of fusion methods}\label{table:1}
\begin{tabularx}{15.9cm}{|p{1.7cm}<{\centering}|p{2.1cm}<{\centering}|p{1.8cm}<{\centering}|p{1.8cm}<{\centering}|p{1.8cm}<{\centering}|p{1.8cm}<{\centering}|p{1.8cm}<{\centering}|}
\hline
\multirow{2}{*}{\textbf{\tabincell{c} {$\!$Measure}}} &
\multicolumn{3}{c|}{$\!$\textbf{Methods in noise-free case}}& \multicolumn{3}{c|}{$\!$\textbf{Methods in noisy case ($\sigma=15$)}}\\
\cline{2-7}
& \textbf{SF-DCT} & \textbf{SR-KSVD} & \textbf{Ours} & \textbf{SF-DCT} 
& \textbf{SR-KSVD} & \textbf{Ours} \\
\hline
\textit{\textbf{$T\!i\!m\!e(\!s\!)$}}
& 1.073 & 4.342 & 4.567  & 2.179 
& 5.281 & 5.981 \\
\hline
\end{tabularx}
\end{table}

\section{Conclusion}
\label{sec:con}
A novel fusion approach for combining multi-focus noisy images into a higher quality all-in-focus image based on analysis sparse model has been presented. Using the cosparsity prior assumption, we have proposed an analysis operator learning approach based on ADMM. Furthermore, an efficient fusion processing via the learned analysis operator has been presented. Extensive experiments have demonstrated that the proposed approach can fuse images with remarkably high-quality, and have confirmed the highly competitive performance of our proposed algorithm. As a future work, a more flexible penalty function can be employed in the fusion problem, which can possibly lead to even better results.

\bibliographystyle{IEEEtran}

\end{document}